\documentclass[conference]{IEEEtran}
\IEEEoverridecommandlockouts
\usepackage{cite}
\usepackage{amsmath,amssymb,amsfonts}
\usepackage{algorithmic}
\usepackage{graphicx}
\usepackage{textcomp}
\usepackage{multirow}
\usepackage{xcolor}
\def\BibTeX{{\rm B\kern-.05em{\sc i\kern-.025em b}\kern-.08em
    T\kern-.1667em\lower.7ex\hbox{E}\kern-.125emX}}
\usepackage{amsmath}
\usepackage{mathtools}
\usepackage{emp}

\begin{document}

\title{Photogrammetry for Digital Twinning Industry 4.0 (I4) Systems\\
\thanks{This work is partly supported by the National Science Foundation (NSF) award number NSF-2335046 and the University of Arizona's Research, Innovation \& Impact (RII) award for the ``Future Factory’’.}
}
\author{
    \IEEEauthorblockN{
        Ahmed Alhamadah\IEEEauthorrefmark{1}, 
        Muntasir Mamun\IEEEauthorrefmark{1},
        Henry Harms\IEEEauthorrefmark{1}, 
        Mathew Redondo\IEEEauthorrefmark{1},
        Yu-Zheng Lin\IEEEauthorrefmark{2},\\
        Jesus Pacheco\IEEEauthorrefmark{3},
        Soheil Salehi\IEEEauthorrefmark{2}, 
        and Pratik Satam\IEEEauthorrefmark{1}
    }
    \IEEEauthorblockA{\IEEEauthorrefmark{1}Department of Systems and Industrial Engineering, University of Arizona\\
    \IEEEauthorrefmark{2}Department of Electrical and Computer Engineering, University of Arizona\\
    \IEEEauthorrefmark{3}School of Industrial Engineering, University of Sonora, Mexico\\
    Email: \{alhamadah, muntasir, henryharms, mredondo245, pratiksatam\}@arizona.edu,\\ yuzhenglin@arizona.edu, ssalehi@arizona.edu, jesus.pacheco@unison.mx}
}

\maketitle

\begin{abstract}
The onset of Industry 4.0 is rapidly transforming the manufacturing world through the integration of cloud computing, machine learning (ML), artificial intelligence (AI), and universal network connectivity, resulting in performance optimization and increase productivity. Digital Twins (DT) are one such transformational technology that leverages software systems to replicate physical process behavior, representing the physical process in a digital environment. This paper aims to explore the use of photogrammetry (which is the process of reconstructing physical objects into virtual 3D models using photographs) and 3D Scanning techniques to create accurate visual representation of the 'Physical Process', to interact with the ML/AI based behavior models. To achieve this, we have used a readily available consumer device, the iPhone 15 Pro, which features stereo vision capabilities, to capture the depth of an Industry 4.0 system. By processing these images using 3D scanning tools, we created a raw 3D model for 3D modeling and rendering software for the creation of a DT model. The paper highlights the reliability of this method by measuring the error rate in between the ground truth (measurements done manually using a tape measure) and the final 3D model created using this method.  The overall mean error is 4.97\% and the overall standard deviation error is 5.54\%  between the ground truth measurements and their photogrammetry counterparts. The results from this work indicate that photogrammetry using consumer-grade devices can be an efficient and cost-efficient approach to creating DTs for smart manufacturing, while the approaches flexibility allows for iterative improvements of the models over time. 

\end{abstract}


\section{Introduction}

As digital technologies have become more accessible and powerful, Industry 4.0 has emerged as the next advancement in manufacturing, development, planning, and education as the digital and physical world continue to intersect \cite{rasheed2020digital}. Advancements in industrial methods that have enabled I4.0 often involve the automation of cyber-physical systems or the processing of many signals to be analyzed in behavioral and physical models of real industrial processes \cite{lin2023dt4i4}. Across many sectors, digital modeling of cyber-physical relationships has allowed for virtual environments that digitally mirror real systems, such that the term \emph{Digital Twin (DT)} has come to describe many different types of systems that all digitally mirror a real system’s behavior \cite{rasheed2020digital,lin2023dt4i4,wright2020tell}. 

A DT is a Software System replicating the behavior of one or more physical process using one or more behavior models, aiming to represent the physical twin's complete lifecycle \cite{lin2023dt4i4}. DT's aim to combine data-based and physics-based behavior models, while driven dynamically by real data, allowing for observation and analysis of processes over time \cite{wright2020tell,qi2021enabling}. The most advanced twins may connect to real-time data, like an interactive 3D simulation of an active offshore oil platform, while others, sometimes described as digital siblings, enable testing hypothetical situations or performing analysis in an accurate, data-driven environment \cite{rasheed2020digital}. This can be particularly useful in manufacturing processes as product design, performance, maintenance, parameters, and assessments can be tested and adjusted in a virtual environment that reacts like the physical system without posing a risk to safety or production \cite{wright2020tell}. Other industries have also found benefits in using DT environments over simple digital models for the ability to customize environments to specific situations. However, 3D modeling relies on manual use of Computer-Aided Design (CAD) software models that are difficult to acquire, proprietary, and unavailable for heterogeneous systems. For example, for an actual deployment of a factory production line, obtaining a 3D model of each machine is expensive, or not possible, posing a challenge in building a \emph{true} DT, requiring an efficient framework for the 3D models of each machine combining them with the DT.

\begin{figure*}[t]
    \centering
    \includegraphics[scale=0.40]{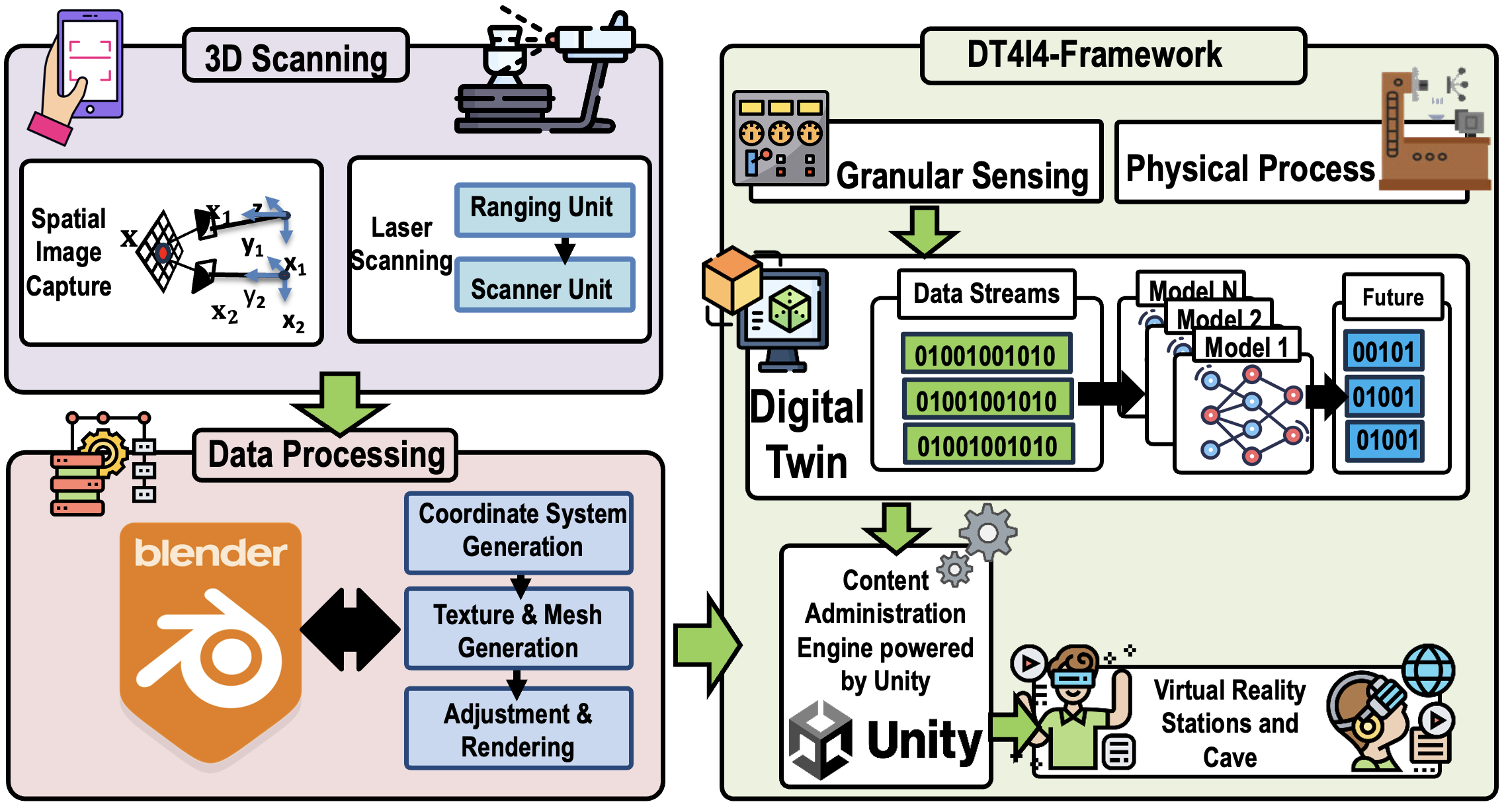}
    \caption{3D Scanning and Reconstruction for Industry 4.0 Digital Twin Framework }
    \label{fig:arch}
\end{figure*}

Through this paper, we establish a methodology to efficiently build 3D models for digital twinning of Industry 4.0 systems through integration into digital twinning frameworks like the DT4I4-Secure framework proposed by Lin et al.\cite{lin2023dt4i4} as shown in figure \ref{fig:arch}, using photogrammetry. The main contributions of this paper:
\begin{itemize}
    \item The paper presents a methodology to produce 3D models for building DTs, at scale, accurately, and at inexpensive cost allowing accurate replication of physical systems in a virtual space. 
\end{itemize}
\begin{itemize}
    \item The paper showcases the use of stereo-vision photogrammetry to create a more accurate 3D reconstruction of the physical system.
\end{itemize}
\begin{itemize}
    \item The paper highlights procedural considerations and error rates in results between the ground truth measurements and photogrammetry model.
\end{itemize}

The rest of the paper is organized as follows: In Section II we discuss literature related DT and photogrammetry. In section III we highlight our methodology to use photogrammetry for building the accurate 3D models. In section IV we discuss experimental evaluation and results. We finally conclude the paper in section V.


\section{Literature Review}

This section presented the related work for this paper. The related work is divided into two subsections: Digital Twin and Photogrammetry.
\subsection{Digital Twin (DT)}
A DT aims to establish a mirrored connection between the physical and virtual realms, mapping sensor-measured data onto the virtual model. NASA’s 2010 technology roadmap draft outlined the utilization of DTs as physical models, updated through sensor feedback to reflect vehicle conditions \cite{glaessgen2012digital}. Tao et al. propose a  DT to be five-dimensional: Physical, Virtual, Connection, Data, and Service. In this five-dimension model, the DT can be applied for several different applications including predictive analysis \cite{shafto2012modeling}, optimization, and security \cite{holmes2021digital} \textcolor. Lin et al. \cite{lin2023dt4i4}, present a similar five-layered framework to DTs to address Industry 4.0 Security challenges. Similar to Lin's work, DTs find application in system design, optimization, predictive analysis, and education requiring accurate representation and visualization of the physical process to improve the DT's usability \cite{rasheed2020digital}. Photogrammetry is an effective method to visualize the physical systems of the DT.

 \subsection{Photogrammetry}
 Photogrammetry is the measurement of an object's distances from multiple photographs of the same object \cite{Magnani2018}. Photogrammetry through stereo-vision detects the depth between images to generate accurate 3d models of physical objects \cite{magnani2020digital}. While traditionally restricted to specialized scanning equipment, the Internet of Things  (IoT) revolution has made photogrammetry accessible through low end/low cost devices and the combination of digital cameras, LiDAR (Light Detection and Ranging) scanners, and software (combined with artificial intelligence) \cite{anderson1982photogrammetry, fussell1982terrestrial,turpin1979stereophotogrammetric}. Photogrammetry methods such as stereo-vision photogrammetry enable highly accurate 3D modeling through multiple cameras capturing complex systems from various angles utilizing two horizontal synchronized cameras \cite{yin2020stereovision,wang2015stereo ,samper2013stereo }, with research efforts focused on reduction in measurement uncertainty through using quaternions modeling of the stereo cameras or use of triangulation \cite{lavecchia2017influence, GUM1995,chen2021analytical}. These advancements allow the usage of stereo-vision photogrammetry in manufacturing \cite{yin2020stereovision, wang2015stereo, samper2013stereo} and robotics \cite{wang2021mobile,tan2011triple,oh2007development}.

\section {Methodology}

Our methodology aims to leverage photogrammetry to obtain 3D models to integrate with the DT4I4 Framework as shown in Figure \ref{fig:arch}. 3D models requiring a reconstruction of images from 2D to 3D, a process that is broken down into multiple stages: Camera Calibration Stage, Image Pair Rectification Stage, and Space Point coordinate calculation Stage. This section describes each of these stages.

\subsection{Camera Calibration}

\begin{figure}[t!]
    \centering
    \includegraphics[width=\linewidth]{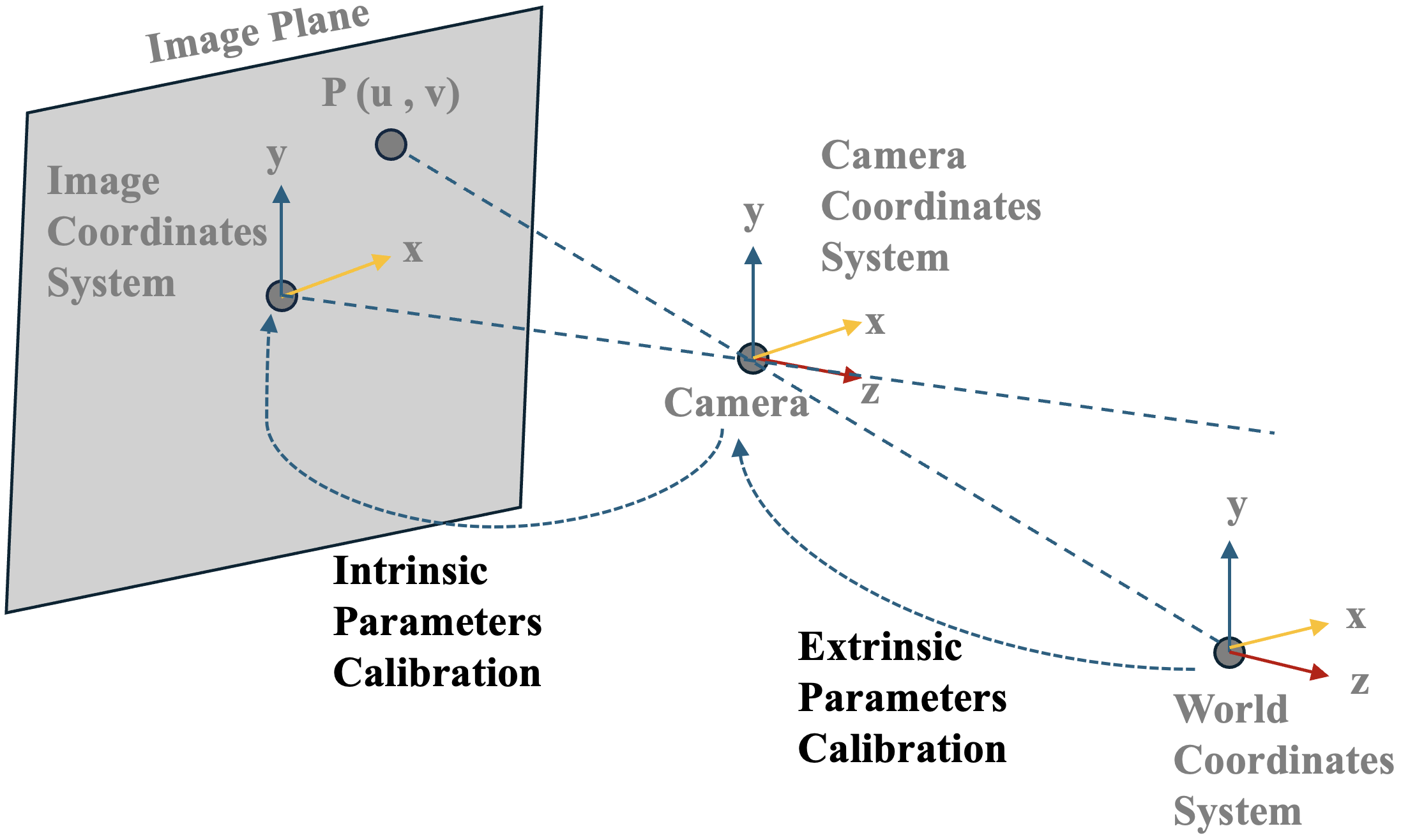}
    \caption{Camera Calibration Process}
    \label{fig:camera vision}
\end{figure}

Camera calibration extracts metric information from 2D images \cite{zhang2004camera},  to estimate a camera's intrinsic and extrinsic parameters, relating the 2D image plane points (in the camera's captured image), to corresponding 3D points in the real world\cite{mathworks_camera_calibration}. Intrinsic parameters describe the camera's internal characteristics like focal length, skew, and optical center, while the extrinsic parameters define the camera's location and orientation in 3D space \cite{heikkila1997four,zhang2004camera} . These intrinsic and extrinsic parameters are estimated by capturing images of a known geometry. This allows us to interpret each sensor pixel as a ray cast into the scene, providing crucial directional information about captured objects. With accurate calibration, we can extract a rigid transformation to convert 3D world coordinates into the camera's 3D coordinates \cite{mathworks_camera_calibration}. This process shown in Figure \ref{fig:camera vision} enables accurate camera calibration allowing accurate reconstruction of the scene's geometry. \cite{bouguet_camera_calibration_toolbox, berger2009geometry}.

 These parameters are configured using a planar pattern-based camera calibration algorithm \cite{zhang2000flexible}. A point on a 2D plane is represented by \( k_s = \begin{matrix} [u \, v] \end{matrix} \). A point on a 3D plane is represented by \( K_s = [X_w, Y_w, Z_w] \). To scale these, 1 is added to the last element of each of them. \( k = [u, v, 1] \) and \( K = [X_w, Y_w, Z_w, 1] \). Through the pinhole phenomena, we gain the relationship between both 3D points and their 2D projection\cite{zhang2000flexible}. Equations (1) and (2) explain that relationship. 

\begin{equation}
    ck_s = b \begin{bmatrix} 
N & e
\end{bmatrix} K_s
\end{equation}

\begin{equation} 
\begin{split}c
\begin{bmatrix}
u\\
v\\
1\\
\end{bmatrix} = \begin{bmatrix}
\alpha_{u} & \gamma & u_{0}\\ 
0 & \alpha_{v} & v_{0}\\ 
0 & 0 & 1\\
\end{bmatrix}
\begin{bmatrix}
N & e\\
\end{bmatrix}
\begin{bmatrix}
X_w\\
Y_w\\
Z_w\\
1\\
\end{bmatrix}\\
\end{split}
\end{equation}\label{equ:first}

 We denoted the scaling factor as c, the pixel focus as \([ \alpha_u, \alpha_v ]\), which are the scaling factors of images in 2D  plane axes, the distortion parameters as \(\gamma\), and the principle point coordinates as \([u_0, v_0]\). The camera's extrinsic parameters include the rotation matrix \(\mathbf{N}\) and the translation vector \(\mathbf{e}\). The relative position of both stereo cameras is determined by calculating the associated external parameters as shown in equation (3): 

\begin{equation}
\left\{
\begin{aligned}
N_{RL} &= N_r N_l^{-1} \\
e_{RL} &= e_r - N_r N_l^{-1} e_l
\end{aligned}
\right.
\end{equation}

The variables $[N_l, e_l]$ and $[N_r, e_r]$ represent the rotation matrices and translation vectors of the stereo cameras on the right and left, respectively, with respect to a specific world coordinate system. Similarly, we denote the rotation matrix and translation vector between the two cameras as \(N_{RL}\) and \(e_{RL}\) respectively.

\subsection{Image Pair Rectification}
Image pair rectification aligns points in the left and right images on the same plane by applying a transmission-projection transformation to rectify the image pairs \cite{fusiello2000compact}. We determined the transmission projection transformation matrices of the left-right pictures and subsequently utilized them in the left-right images using a bilinear interpolation algorithm \cite{gonzalez2008digital}, which interpolates both left and right image variables using linear interpolation throughout this procedure. The projection points in the image pairs must adhere to this fundamental equation :

\begin{equation}
\mathbf{p}_r^T  \mathbf{b}_r^{-T} [\mathbf{e}_{RL}]_x \mathbf{N}_{RL}  \mathbf{b}_l^T \mathbf{p}_l = 0
\end{equation}

 We denoted the left and right projection points of a point by \(\mathbf{p}_l\) and \(\mathbf{p}_N\) respectively. We also denoted the intrinsic parameters of the right and left cameras by \(\mathbf{b}_r\) and \(\mathbf{b}_l\). Similarly, we denoted the antisymmetric matrix by \([\mathbf{e}_{RL}]_x\) which we derived from the translation vector \(\mathbf{e}_{RL}\).

\subsection{Space Point Coordinates Calculation}
We determine the coordinates of points in space using the triangle measuring method,
which we defined using equation (5):

\begin{equation}
\left\{
\begin{aligned}
X_w (x,y) &= \frac{G \times O \times (x_l - u_o l)}{\alpha_{ul} \times (x_l - x_r)} \\
Y_w (x,y) &= \frac{G \times O \times (y_l - v_o l)}{\alpha_{vl} \times (x_l - x_r)} \\
Z_w (x,y) &= \frac{G \times O}{x_l - x_r}
\end{aligned}
\right.
\end{equation}

We denoted the variables as follows: the coordinates of spatial points as  \((X_w, Y_w, Z_w)\); the pixel focal length of the left camera as represented as \([\alpha_{ul}, \alpha_{vl}]\); the optical center distance between both cameras as  \(G\); the pixel focal length of the camera is represented as \(O\); the principal point coordinate of the left camera as \((u_{0l}, v_{0l})\); and \((x_l, y_l)\) and \((x_r, y_r)\) are the rectified coordinates of the left-right projection points \(m_l\) and \(m_r\), respectively.

\begin{table}[b!]
  \centering
  \caption{Camera Specifications}
  \label{tab:camera_specs}
  \begin{tabular}{|l|l|}
    \hline
    \textbf{Camera}          & \textbf{Specification}          \\
    \hline
    & 24 mm                           \\
                             48MP Main& f/1.78                          \\
                             & Sensor-Shift OIS                \\
                             & 24MP/48MP Photos                \\
    \hline
    & 13 mm                           \\
                             12MP Ultra Wide& f/2.2                           \\
                             & $120^\circ$ FOV                  \\
                             & 100 Focus Pixel                 \\
    \hline
  \end{tabular}
  \vspace{1em} 
\end{table}

Consequently, by using a triangulation algorithm, we can fit the spatial point cloud, which are points in space that represent the 3D shape, to obtain a curved surface after we computed it using Equation (5) completing the 3-D reconstruction.
\begin{figure*}[ht]
  \centering
    \includegraphics[width=0.55\linewidth]{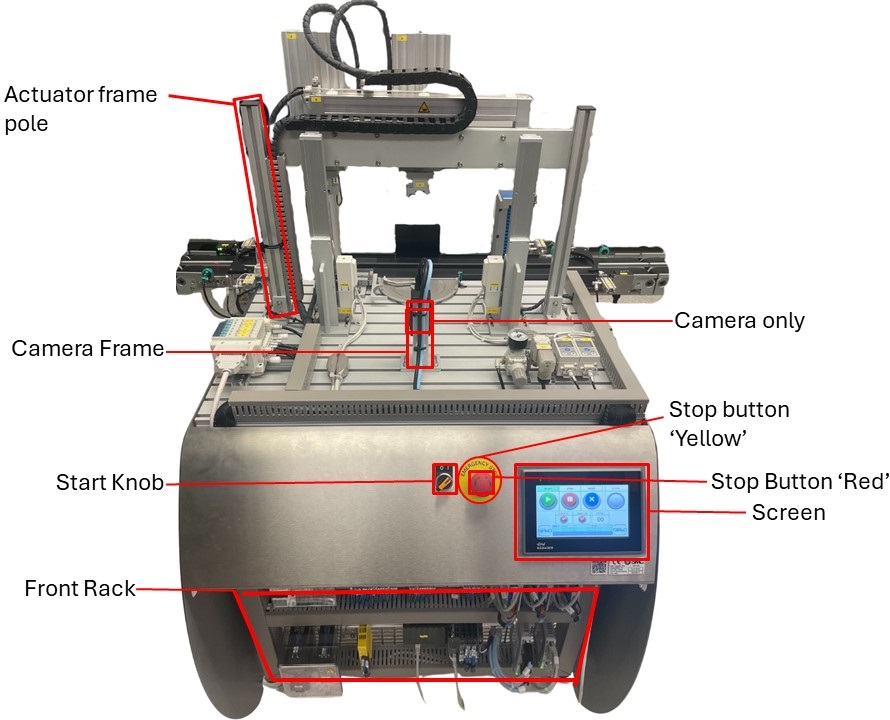}
  \hfill
    \includegraphics[width=0.375\linewidth]{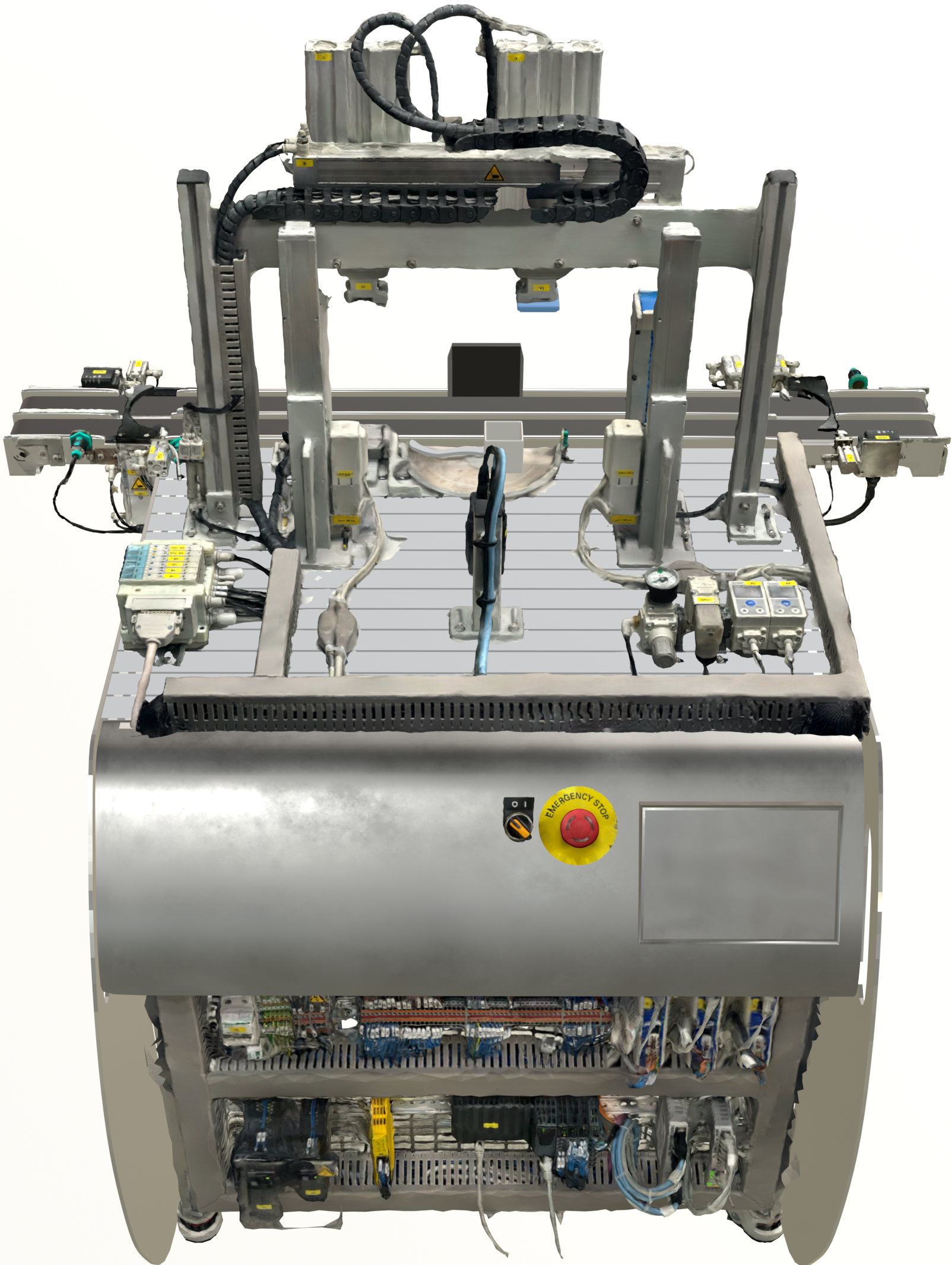}
  \caption{\textbf{A.} Labeled snapshot for the front section, \textbf{B.} Reconstructed 3D model of the front section}
  \label{fig:two-images}
\end{figure*}

\begin{figure*}[h!]
  \centering
  \includegraphics[width=0.50\linewidth]{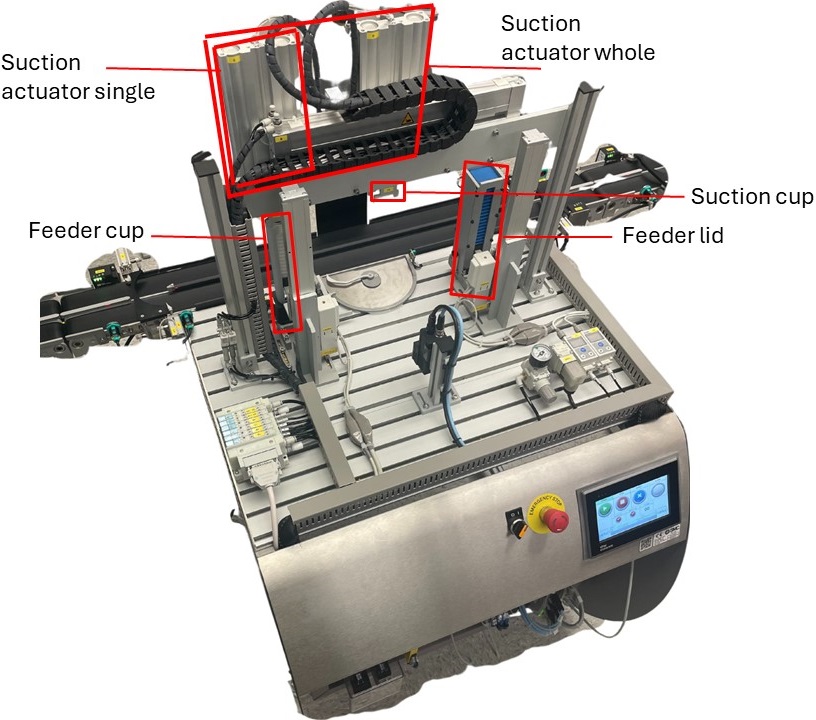}
  \hfill
  \includegraphics[width=0.4\linewidth]{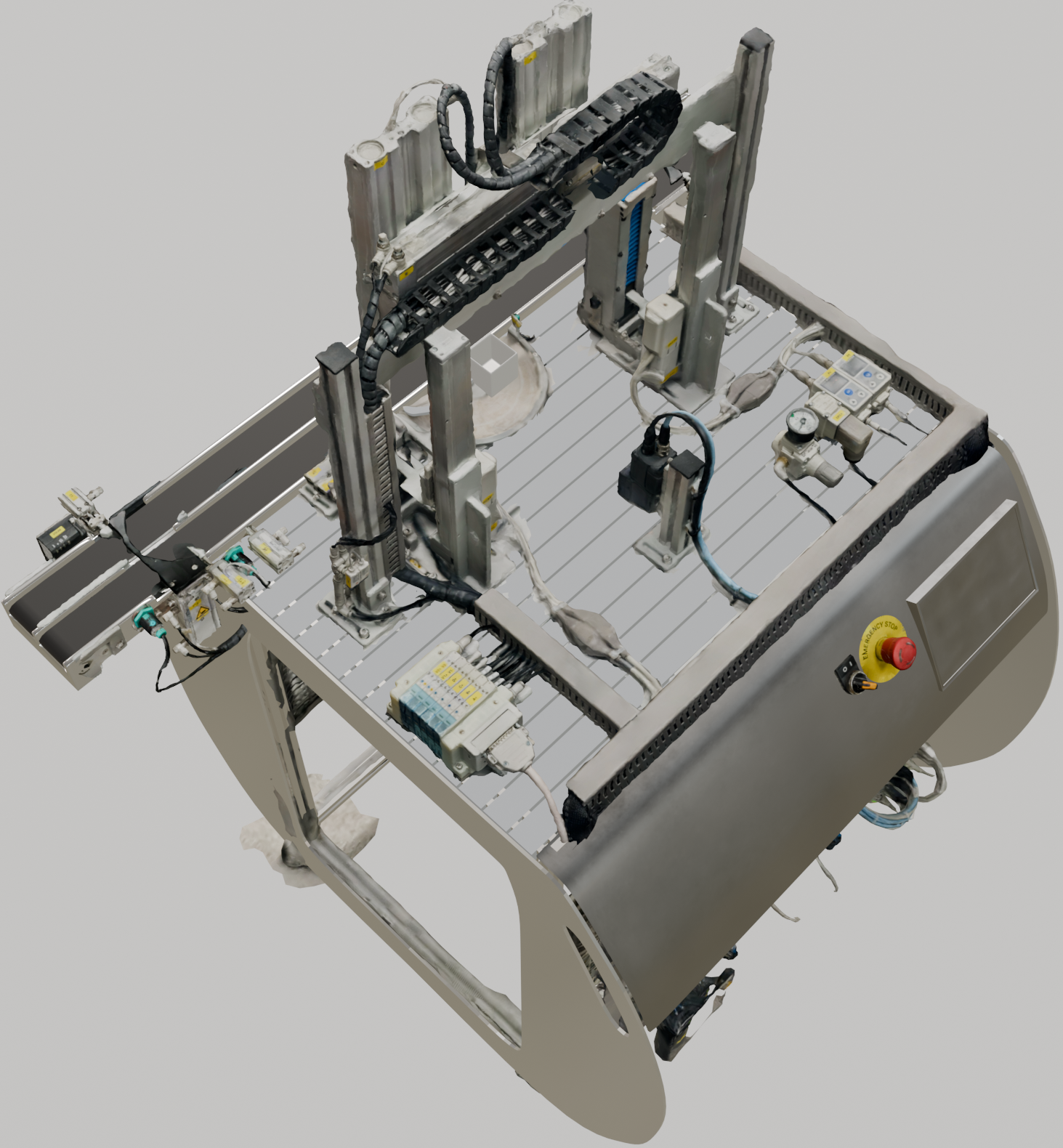}
  \caption{\textbf{A.} Labeled snapshot for the top section, \textbf{B.} Reconstructed 3D model of the top section}
  \label{fig:top}
\end{figure*}

\begin{figure*}[h!]
  \centering
  \includegraphics[width=0.55\linewidth]{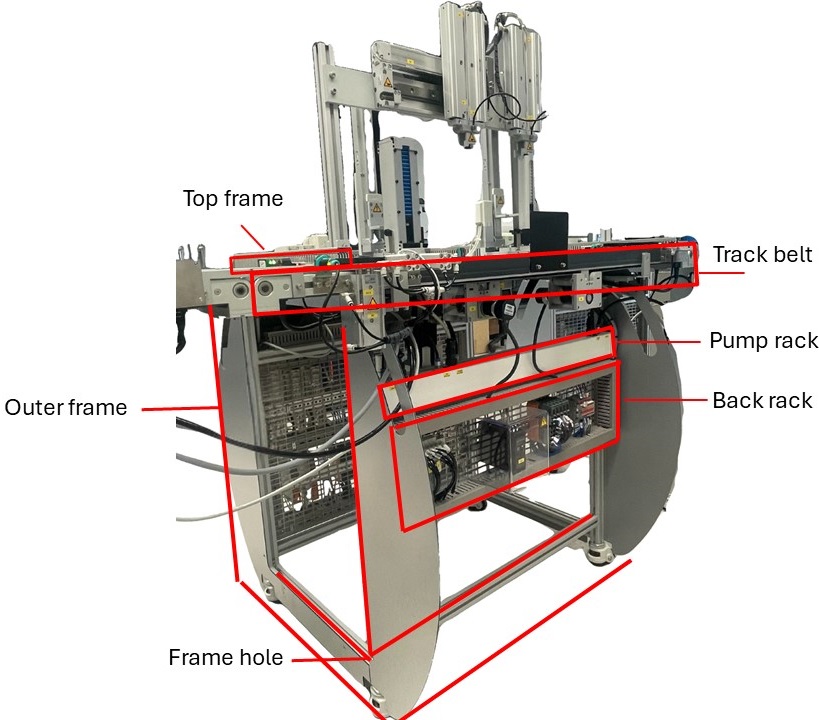}
  \hfill
  \includegraphics[width=0.35\linewidth]{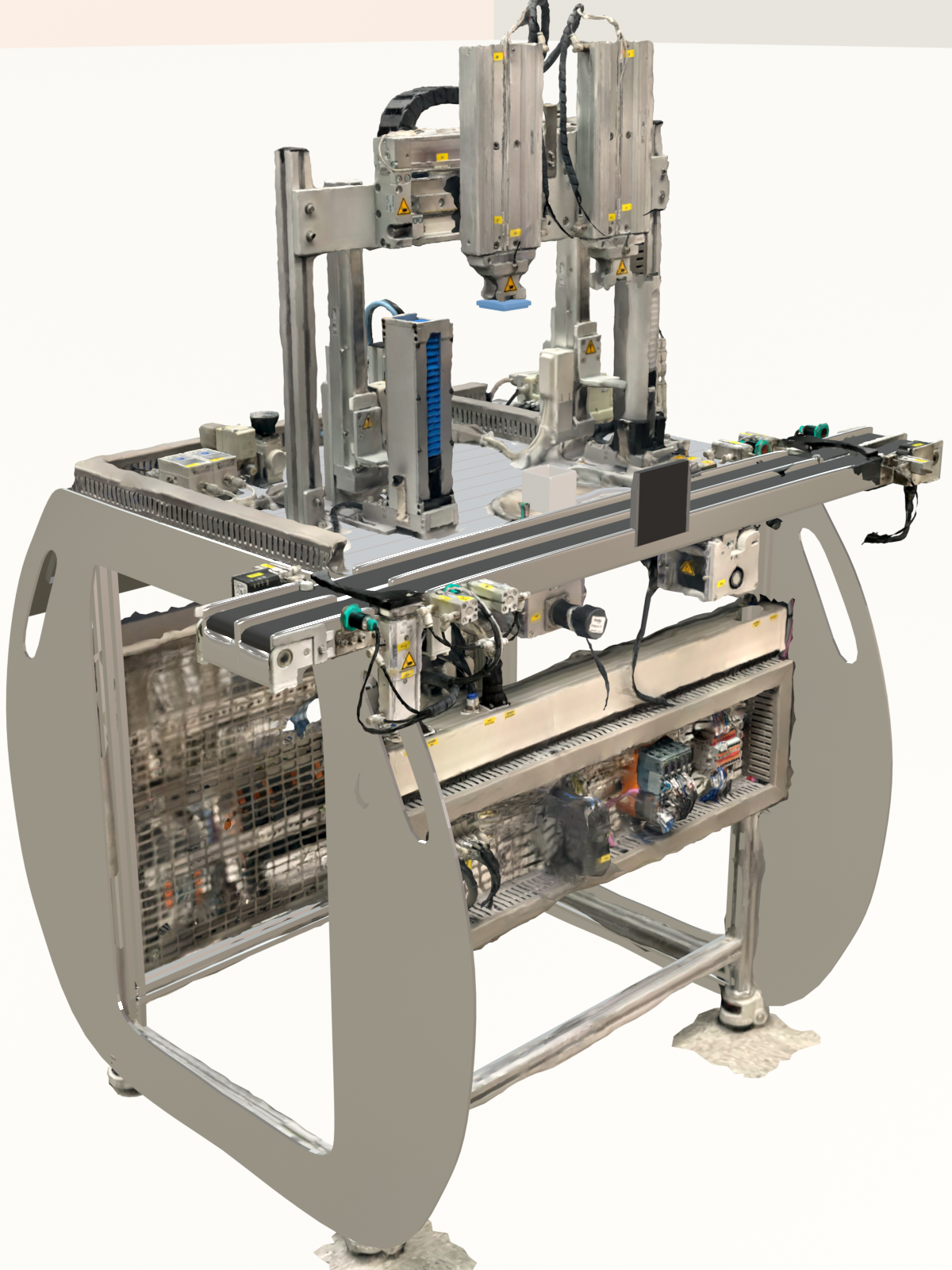}
  \caption{\textbf{A.} Labeled snapshot for the back section, \textbf{B.} Reconstructed 3D model of the back section}
  \label{fig:back}
\end{figure*}

\section{Experimental Evaluations}

\subsection{Experiment Setup}

The section provides an overview of the setup we used to execute this experiment.
\subsubsection{UArizona Future Factory}
The UArizona Future Factory is an Industry 4.0 system comprising of four SMC's Smart Innovation Factory (SIF) 400 stations, developer stations, data collection and historian stations, and attacker station. For the goals of this paper, we use the SIF 400 stations for the 3D modeling. SIF 400 simulates an automated smart factory that utilizes Industry 4.0 technologies, leveraging manufacturing concepts and the connectivity of real-world factories in areas such as production, assembly, logistics, and management. The system is connected machine-to-machine using management software that allows for the process to be fully automated. The SIF-400 at UArizona is made of 4 modular stations that interconnect with a conveyor belt: \textbf{\textit{1) SIF 401: Pallet and Container Feeding Station:}} This station stores containers of two different formats: cylindrical and/or quadrangular prisms, as well as pallets with an integrated RFID “tag”. The containers are placed onto the pallet in order to be transported along the process. \textbf{\textit{2) SIF 402: Container Filling Station - Solids:}} This station fills containers of two different formats: cylindrical and/or quadrangular prisms, with solid prime material. \textbf{\textit{3) SIF 405: Capping Station:}} This station supplies and attaches two different types of cap to containers of two different formats: cylindrical and/or quadrangular prisms. \textbf{\textit{4) SIF 407: Container Labeling and Dispatching Station:}} This station carries out two independent processes: on one hand, labelling the container with a QR matrix code and, on the other hand, dispatching the container to a shipping platform. For the scope of this paper, we focus on evaluating our methodology for building 3D models of the \emph{SIF 405: Capping Station}.

\subsubsection{Stereo-Vision Cameras iPhone 15 Pro}
The iPhone 15 Pro equipped with three cameras and a lidar scanner managed by the latest Apple chips which are A17 Pro chip, a 6‑core CPU with 2 performance and 4 efficiency cores, a 6‑core GPU, and a 16‑core neural engine, is a strong (yet cost effective) platform for performing photogrammetry. The specifications of each camera are highlighted in Table  \ref{tab:camera_specs}. For this work we used stereo images and lidar scans from an iPhone 15 Pro stitched together with AI-assisted tools like Polycam to create 3D models in Blender.

\begin{figure*}[h!]
  \centering
  \includegraphics[width=0.4\linewidth]{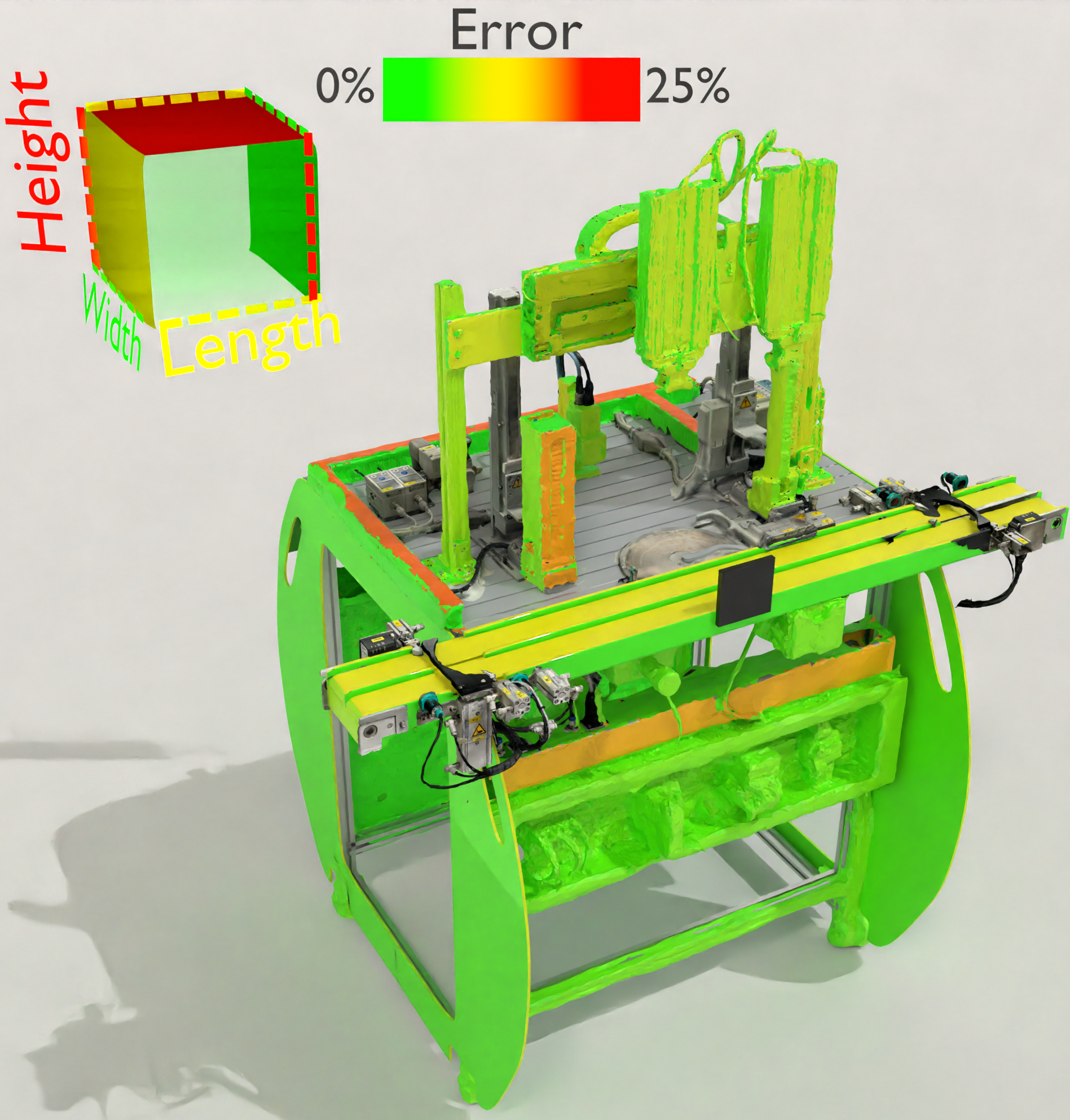}
  \hfill
  \includegraphics[width=0.55\linewidth]{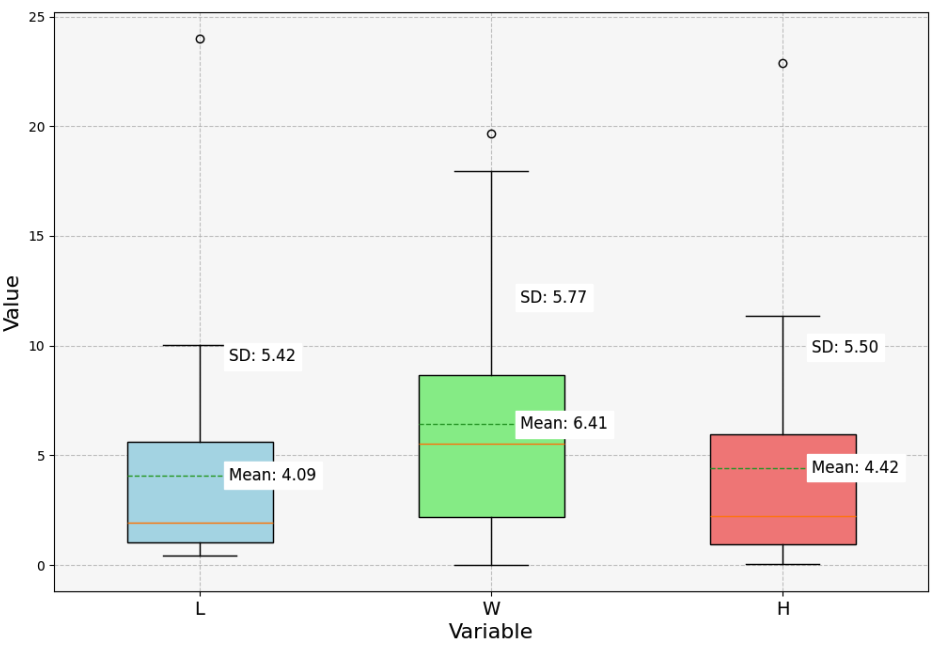}
  \caption{\textbf{A.} Error heatmap, \textbf{B.}       Mean average and standard deviation distribution}
  \label{fig:error}
\end{figure*}

\subsubsection{Software for model creation}
Polycam is an AI-assisted took that enables the stitching of stereo images, and combining them with lidar scans, assisting in building accurate 3D models. Polycam Uses a combination of lidar, triangulation algorithm, and light detection to generate precise 3D point cloud images. This versatility makes Polycam a valuable and accessible tool for creating precise visual digital twins using an iPhone and the app. Blender is an opensource software toolset for transforming raw 3D scans from a scanning application such as Polyam into usable and optimized 3D models, by creating meshes. Additionally, it allows for retopology to create a more efficient and well-structured version of the model by reconstructing the polygonal mesh. Blender also supports texturing and UV unwrapping which enables the application of captured textures (colors and details) onto the 3D model. These provide a more optimized and refined final version of the model.

\begin{table*}[ht]
\centering
\caption{Front Section measurements}
\label{tab:front}
\begin{tabular}{|l|c|c|c|c|c|c|c|c|c|}
\hline
\multirow{2}{*}{\textbf{Objects}} & \multicolumn{3}{c|}{\textbf{L}}                 & \multicolumn{3}{c|}{\textbf{W}}                 & \multicolumn{3}{c|}{\textbf{H}}                 \\ \cline{2-10} 
                                   & \textbf{Actual} & \textbf{iPhone 15 Pro} & \textbf{Error} & \textbf{Actual} & \textbf{iPhone 15 Pro} & \textbf{Error} & \textbf{Actual} & \textbf{iPhone 15 Pro} & \textbf{Error} \\ \hline
Actuator Frame Pole               & 1.125           & 1.193              & 6.04\%          & 1.125           & 1.184              & 5.24\%          & 16.84375        & 16.687             & 0.93\%          \\ \hline
Camera Frame                      & 1.125           & 1.012              & 10.04\%         & 1.125           & 1.069              & 4.98\%          & 6.375           & 6.318              & 0.89\%          \\ \hline
Camera Only                       & 1.75            & 1.706              & 2.51\%          & 2.0625          & 1.943              & 5.79\%          & 3.5             & 3.389              & 3.17\%          \\ \hline
Start Knob (Red Circle)           & 1.125           & 1.395              & 24.00\%         & 1.63            & 1.766              & 8.34\%          & 1.625           & 1.451              & 10.71\%         \\ \hline
Stop Button (Yellow Circle)       & 2.6875          & 2.895              & 7.72\%          & 0               & 0                  & 0\%             & 2.75            & 2.934              & 6.69\%          \\ \hline
Start Switch (O/1)                & 1.125           & 1.133              & 0.71\%          & 1.07            & 1.042              & 2.62\%          & 1.75            & 1.795              & 2.57\%          \\ \hline
Screen                            & 7.875           & 7.556              & 4.05\%          & 0.25            & 0.248              & 0.80\%          & 5.5             & 5.21               & 5.27\%          \\ \hline
Front Rack                        & 29.155          & 28.462             & 2.38\%          & 2.34375         & 2.336              & 0.33\%          & 20.65625        & 20.338             & 1.54\%          \\ \hline
\end{tabular}
\end{table*}

\begin{table*}[ht!]
\centering
\caption{Top section measurements}
\label{tab:top}
\begin{tabular}{|l|l|l|l|l|l|l|l|l|l|}
\hline
\multirow{2}{*}{\textbf{Objects}} & \multicolumn{3}{c|}{\textbf{L}}                 & \multicolumn{3}{c|}{\textbf{W}}                 & \multicolumn{3}{c|}{\textbf{H}}                 \\ \cline{2-10} 
                                   & \textbf{Actual} & \textbf{iPhone 15 Pro} & \textbf{Error} & \textbf{Actual} & \textbf{iPhone 15 Pro} & \textbf{Error} & \textbf{Actual} & \textbf{iPhone 15 Pro} & \textbf{Error} \\ \hline
Suction Actuator Whole         & 12.15625            & 12.056                 & 0.82\%             & 1.21875              & 1.311                 & 7.57\%             & 9.125                & 9.072                 & 0.58\%             \\ \hline
Suction Actuator Single        & 4.25                & 4.306                  & 1.32\%             & 1.21875              & 1.311                 & 7.57\%             & 9.125                & 9.072                 & 0.58\%             \\ \hline
Feeder Lid                     & 2.25                & 2.332                  & 3.64\%             & 2.09375              & 2.506                 & 19.69\%            & 9.59375              & 9.588                 & 0.06\%             \\ \hline
Feeder Cup                     & 2.125               & 2.254                  & 6.07\%             & 1.5                  & 1.456                 & 2.93\%             & 10.375               & 9.593                 & 7.54\%             \\ \hline
Suction Cups                   & 3.4375              & 3.46                   & 0.65\%             & 1.375                & 1.468                 & 6.76\%             & 2.125                & 2.186                 & 2.87\%             \\ \hline
\end{tabular}
\end{table*}

\subsection{Experiments}

The process we followed to perform the photogrammetry experiment to create a visual digital twin of the SIF 400 consists of four steps: The first step we followed was to obtain the ground truth measurement. We used the tape measure to manually measure every important dimension (including height, length, and width measurements) of the testbed and recorded the results. The second step we followed was to acquire the images. We leveraged stereo-vision photogrammetry technology using the iPhone’s 15 Pro cameras to create raw 3D images of the testbed. The next step we followed was processing the images using Polycam. We uploaded the images taken into Polycam cloud database and chose the “full” setting which provides detailed physically based rendering (PBR) texture maps and "Raw" to obtain a single texture file. We also used Polycam to perform object masking to help with the environment surrounding the object in the images such as shiny surfaces. The final step we followed was we digitally processed Polycam output images using Blender. We used its manipulation features to adjust the position, scale, and orientation of the 3D model. It helped us to take precise measurements from the images to compare with the ground truth we obtained from the first step.

\begin{table*}[ht]
\centering
\caption{Back section measurements}
\label{tab:back}
\begin{tabular}{|l|l|l|l|l|l|l|l|l|l|}
\hline
\multirow{2}{*}{\textbf{Objects}} & \multicolumn{3}{c|}{\textbf{L}}                 & \multicolumn{3}{c|}{\textbf{W}}                 & \multicolumn{3}{c|}{\textbf{H}}                 \\ \cline{2-10} 
                                   & \textbf{Actual} & \textbf{iPhone 15 Pro} & \textbf{Error} & \textbf{Actual} & \textbf{iPhone 15 Pro} & \textbf{Error} & \textbf{Actual} & \textbf{iPhone 15 Pro} & \textbf{Error} \\ \hline
Top Frame                 & 28.5                & 28.303                & 0.69\%             & 20                  & 20.003                & 0.02\%             & 2.47                 & 1.905                 & 22.87\%            \\ \hline
Frame Hole                & 31.5                & 31.151                & 1.11\%             & 21.97               & 18.728                & 14.76\%            & 28.75                & 28.452                & 1.04\%             \\ \hline
Track Belt                & 48.375              & 45.73                 & 5.47\%             & 5.25                & 5.216                 & 0.65\%             & 2.31                 & 2.048                 & 11.34\%            \\ \hline
Pump Rack (White)         & 29.875              & 29.471                & 1.35\%             & 2.78125             & 2.282                 & 17.95\%            & 2.34375              & 2.313                 & 1.31\%             \\ \hline
Back Rack                 & 29.375              & 28.929                & 1.52\%             & 2.34375             & 2.276                 & 2.89\%             & 8.34375              & 8.188                 & 1.87\%             \\ \hline
Outer Frame               & 31.75               & 31.364                & 1.22\%             & 41.47               & 37.501                & 9.57\%             & 34.25                & 32.294                & 5.71\%             \\ \hline
\end{tabular}
\end{table*}

\subsubsection{Experiment 1: Measuring the accuracy of the front section}

In our first experiment, we recorded the length (L), width (W), and height (H) of major components from the front view of the SIF 405: Capping Station. These measured dimensions offered a baseline set of data to compare with scanned data later. After capturing lidar and photogrammetry scans, we found that while lidar imagery imported to Blender with real-world measurement units, photogrammetry scans had relative origins, rotations, and scales. Blender allowed for aligning the photogrammetry scans to the lidar scans dimensionally and by texture details. This enabled measuring the final 3D model in real-world units to compare them with the baseline data. Figure \ref{tab:front} shows a labeled snapshot of the capping station's front view and the modeled front view to compare it with.

We compared the measured dimensions from the scanned reconstructions to the actual measurements and found error rates between measurements as shown in \ref{tab:front}. We noted that some dimensions saw higher error rates in the scanned data, 10-20\%, particularly for the small details of I/O buttons at the front of the testbed. The imaging process was limited to the physical geometry of the iPhone, so thin details could be obscured in close up imaging producing not fully accurate 3D reconstruction. However, most scanned dimensions were accurate within 5-10\% to the actual measurements, with some errors like the width of the front rack, height of the actuator frame pole, and length of the start switch having under 1\% error.

\subsubsection{Experiment 2: Measuring the accuracy of the top section}

In our second experiment, we recorded the length (L), width (W), and height (H) of major components from a side view of the testbed station. Figure shows \ref{fig:top} , shows the modeled front view compared with a labeled snapshot of the capping station's side view. We took the data obtained from both photogrammetry and ground truth and provided their measurements and corresponding error rates in \ref{tab:top}. In This experiment we focused on larger components that made up dominant volumes of the full testbed like the outer frame of the testbed, the interior hole of the frame, and the track belt. We observed that most dimensions were within an error range of 1-6\%. However, some measurements in the W direction suffered higher rates of error around 10-18\%, and we attributed this to environmental conditions in the physical alignment of the testbed station alongside other stations. We also found out that some H measurements, such as for the top frame and track belt, also had similarly high error rates due to the complexity of the background increasing difficulty for the photogrammetry algorithm to distinguish the edges of these objects.

\subsubsection{Experiment 3: Measuring the accuracy of the back section}
For the final experiment, we recorded the actual measurements of length (L), width (W), and height (H) of major components from a top view of the station before imaging, processing, and digital measurements. Figure \ref{fig:back}, shows the modeled front view compared with a labled snapshot of the capping station's back view. We included the actual measurements, scanned measurements, and their error rates in Table \ref{tab:back}. For this set of dimensions, most parts saw error rates of scanned dimensions under 8\%, with many reaching below 1\% error. Similar to the previous experiment, some measurements had higher errors in the W direction as more geometry was obscured by nearby stations. We noted that the W measurement of the feeder lid part had a disproportionately high error rate of over 19\%, indicating that the physical shape of the photogrammetry camera can affect ability to maneuver around tight spaces and collect fully accurate spatial data. We highlighted the error distribution and focus in the heat map Figure \ref{fig:error}  showing where the errors mostly occurred in the 3D model as we discussed in the experiments. We also provide the overall mean and standard deviation values of the error in figure \ref{fig:error}.

\section{conclusion}

In conclusion, this experiment considered the importance of accurately reflecting physical systems in virtual spaces for the services and benefits produced by Digital Twins as a transformational technology for Industry 4.0 systems. We presented a methodology of producing visualizations for Smart Manufacturing Digital Twins by leveraging consumer-grade hardware and software that was capable of reproducing most real dimensions accurately in virtual space. This method relied on the capabilities of the iPhone 15 Pro, IoT and networked scanning services, and 3D processing software to reliably replicate a physical system in a 3D model such that physical processes may be rendered, driven, and interacted with in a Digital Twin environment. Our experiments proved the reliability of simple photogrammetry methods by comparing digital measurements of the dimensions of critical components to actual measurements and found an overall mean error of 4.97\% with a standard deviation of 5.54\%. Some factors lead to higher dimensional error rates, including subject location relative to other objects, complexity of the background and other environmental conditions, and tight spaces around parts of the subject. These variables created conditions that could skew the virtual representation of certain physical traits, however, the method still produced roughly 90-95\% accurate dimensions. In most cases, we found this method to produce visual results effective enough to enable reflecting of physical processes in a cyber-physical system's relationship.

\section*{Acknowledgment}
This work is partly supported by the National Science Foundation (NSF) award number NSF-2335046 and the University of Arizona's Research, Innovation \& Impact (RII) award for the ``Future Factory’’.

\clearpage \bibliography{refs.bib}
\bibliographystyle{IEEEtran}

\end{document}